\documentclass[letterpaper, 10 pt, conference]{ieeeconf}  

\IEEEoverridecommandlockouts                              

\usepackage{hyperref}
\usepackage{multicol}
\usepackage{multirow}
\usepackage{tabularx}
\usepackage{array}
\newcolumntype{$}{>{\global\let\currentrowstyle\relax}}

\newcolumntype{^}{>{\currentrowstyle}}

\usepackage{subcaption}
\usepackage[table]{xcolor}

\usepackage[american]{babel}
\def\CC{{C\nolinebreak[4]\hspace{-.05em}\raisebox{0.1ex}{\bf ++}}}

\usepackage{graphics} 
\usepackage{epsfig} 
\usepackage{mathptmx} 
\usepackage{times} 
\usepackage{amsmath} 
\usepackage{amssymb}  
\usepackage{marvosym,listings,etoolbox}

\graphicspath{{Figures/}}

\newcommand{\name}{\mbox{LiDAR-Flow}}

\title{\LARGE \bf \name{}: Dense Scene Flow Estimation from Sparse LiDAR and Stereo Images}

\author{Ramy Battrawy$^{1}$, Ren{\'e} Schuster$^{1}$, Oliver Wasenm{\"u}ller$^{1}$, Qing Rao$^{2}$, Didier Stricker$^{1}$
\thanks{$^{1}$DFKI -- German Research Center for Artificial Intelligence, Germany:
	{\tt\small firstname.lastname@dfki.de}}%
\thanks{$^{2}$BMW Group, Germany:
        {\tt\small firstname.lastname@bmw.de}}%
}
\begin{document}

	\maketitle
	\thispagestyle{empty}
	\pagestyle{empty}

	\begin{abstract}
		We propose a new approach called \name{} to robustly estimate a dense scene flow by fusing a sparse LiDAR with stereo images. We take the advantage of the high accuracy of LiDAR to resolve the lack of information in some regions of stereo images due to textureless objects, shadows, ill-conditioned light environment and many more. Additionally, this fusion can overcome the difficulty of matching unstructured 3D points between LiDAR-only scans. Our \name{} approach consists of three main steps; each of them exploits LiDAR measurements. First, we build strong seeds from LiDAR to enhance the robustness of matches between stereo images. The imagery part seeks the motion matches and increases the density of scene flow estimation. Then, a consistency check employs LiDAR seeds to remove the possible mismatches. Finally, LiDAR measurements constraint the edge-preserving interpolation method to fill the remaining gaps. 
		In our evaluation we investigate the individual processing steps of our \name{} approach and demonstrate the superior performance compared to image-only approach.
	\end{abstract}

	\section{INTRODUCTION}
		Robust perception is an essential task for reliable autonomous driving platforms. Achieving this goal requires more awareness  of the dynamic changes of the environment. For this purpose, dense scene flow estimation became an active research area. It seeks to compute the 3D geometry as well as the 3D motion field and serves as comprehensive representation of a dynamic environment. In the past, this was often approximated by optical flow in 2D space.
		
		Scene flow is usually computed based on dense pixel matches in stereo images.
		Dense pixel matching is well established \cite{menze2015object, vcech2011scene, wedel2011stereoscopic} and achieves reasonable quality in many scenarios.
		However, in image regions with textureless objects, shadows, ill-conditioned light environment, etc., image-based matching is extremely challenging.
		However, exactly these regions are of high importance for autonomous vehicles and inaccuracies can anticipate accidents scenarios.
			
		Unlike image-based technologies, LiDAR sensors are much less sensitive to the aforementioned environmental conditions.
		Thus, LiDAR is a core technology for mapping 3D surroundings of autonomous vehicles.
		However, dense LiDAR sensors are expensive and matching unstructured 3D point clouds is challenging \cite{thomas2019delio}.
		\begin{figure}[t]
			\centering
			\begin{subfigure}[h]{1.0\columnwidth}  
				\includegraphics[scale=1.0]{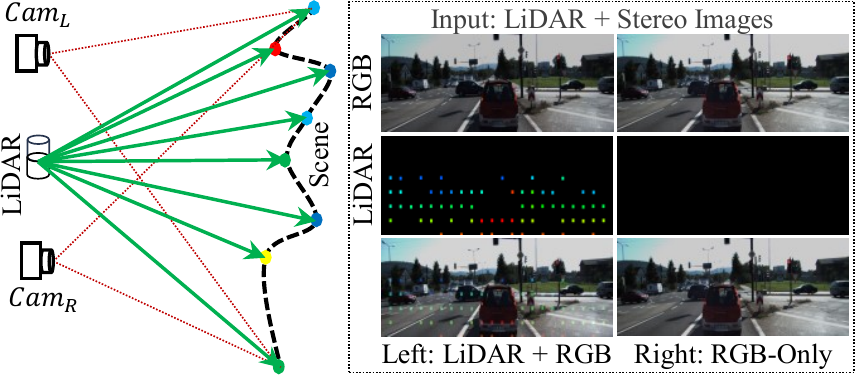}
				\caption{\name{}: Sparse LiDAR fused with stereo images}	
			\end{subfigure}
			\begin{subfigure}[h]{0.49\columnwidth} 
				\includegraphics[scale=1.0]{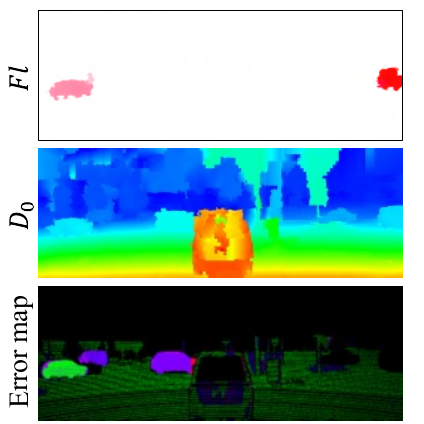}%
				\caption{Image-only scene flow \cite{schuster2019SceneFlowFieldsPlusPlus}}
			\end{subfigure}
			\begin{subfigure}[h]{0.49\columnwidth}
			\includegraphics[scale=1.0]{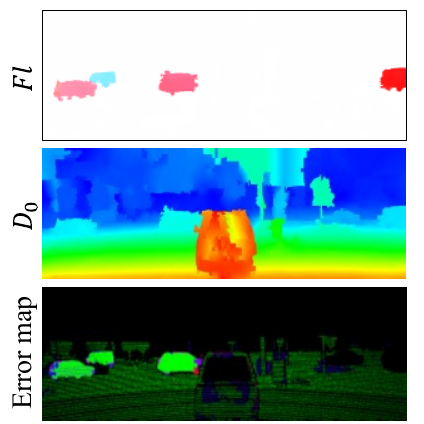}
			\caption{Our \name{}}
			\end{subfigure}
		
			\caption{We introduce \name{} which fuses a sparse LiDAR and stereo images for dense scene flow. Compared to image-only approach \cite{schuster2019SceneFlowFieldsPlusPlus}, we achieve high robustness in challenging image regions with our fusion. The green pixels of the error map represent the inliers of scene flow compared to the ground truth.
			}	
			\label{Fig1: Introduction}
		\end{figure}	
		Our approach fuses sparse LiDAR values and stereo images (see Fig. \ref{Fig1: Introduction}) in order to overcome the deficiencies of each technology which leads to increase in the robustness of scene flow estimation. 
		To the best of our knowledge, \name{} is the first attempt for this fusion in the context of scene flow.
		
\begin{figure*}[t]
			\centering
			\includegraphics[scale=1.0, width = \textwidth ]{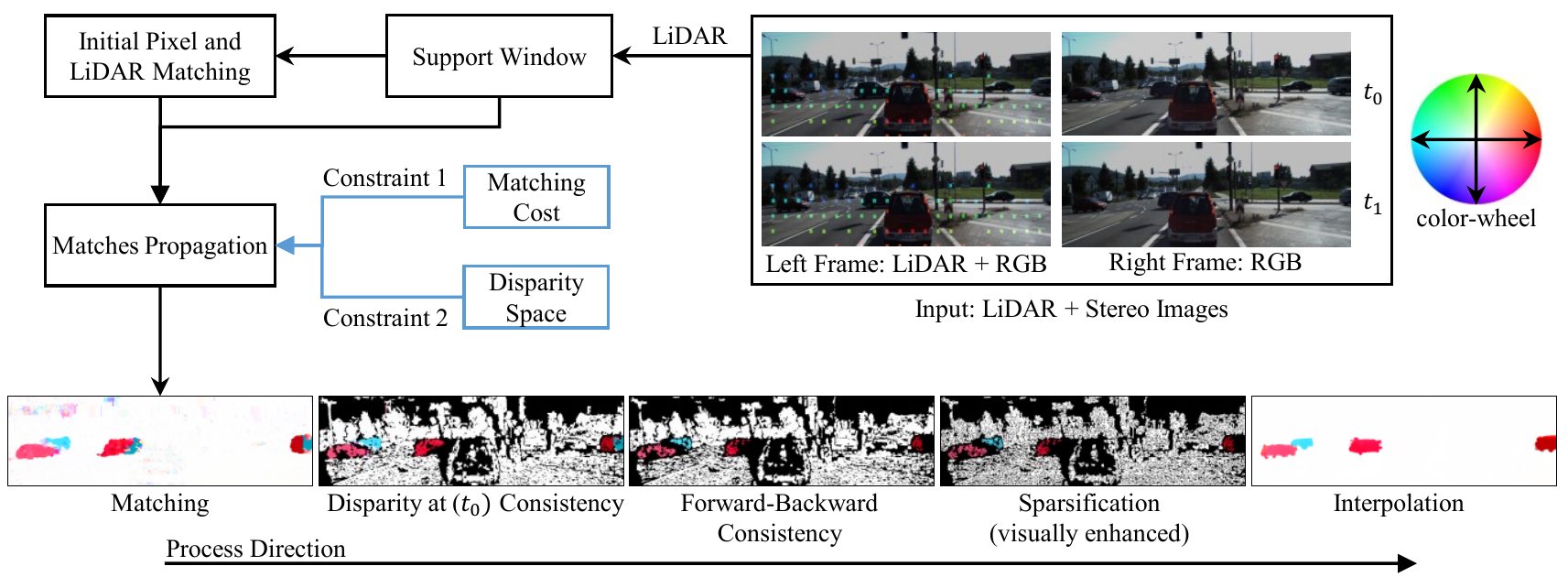}
			\caption{Overview of our \name{}. We show the optical flow $(Fl)$ estimation after each step. The color-wheel encodes the motion direction and the black color maps the removed outliers after consistency check and sparsification.}
			\label{Fig2:Overview}
		\end{figure*}

		Our \name{} utilizes LiDAR measurements as anchor points to support each processing step across the framework.
		We utilize a sparse-to-dense approach for our \name{}, which is established for image-only scene flow already \cite{schuster2019SceneFlowFieldsPlusPlus, hu2017robust,hu2016efficient,schuster2018sceneflowfields}.
		We initialize disparity with the LiDAR measurements, estimate flow based on image matching and iteratively propagate these measurements with random search into the neighborhood. 
		During propagation process, each scene flow pixel will be optimized by minimizing a matching cost function. 
		Mismatches  will be removed through consistency check with the help of LiDAR measurements afterwards. 
		The sparse set of matches is robustly interpolated with the LiDAR-support. Fig. \ref{Fig2:Overview} illustrates an overview of our \name{} approach.
		
		In a summary, our contributions are the following:
		\begin{itemize}
			\item We propose \name{} -- the fusion of sparse LiDAR and stereo images for scene flow estimation.
			\item Using LiDAR measurements, we increase the constancy of pixel-based matches for untextured regions.		
			\item We propose an edge-preserving interpolation constrained by LiDAR.
			\item We show the impact of \name{} in each step and compared to the state-of-the-art image-based scene flow on KITTI.
		\end{itemize}

	\section{RELATED WORK}
	The initial approaches for dense scene flow were based on a stereo camera setup.
	Vedula et al. \cite{vedula1999three} combine optical flow estimation with the first-order approximations of depth maps as variational optimization framework. 
	Wedel et al. \cite{wedel2008efficient} use stereo images for scene flow estimation and decouple the motion from geometry for real time use. In contrary, other methods \cite{li2008multi, basha2013multi} couple the motion and geometry consistency to improve the scene flow accuracy under stereo use. In the same context, \cite{ vcech2011scene,hung2013consistent} add the cross constancy and an assumption of gradient constancy  \cite{hung2013consistent,huguet2007variational} increase the robustness of matches against illumination change. Further contribution of \cite{vogel20113d} assumes local rigid regularization instead of variational regularization in using stereo images. In that way, a method of Menze et al. \cite{menze2015object} segments the object scene into a set of rigid objects and uses CRF to estimate the association between them. That method does not consider the non-rigidly moving objects like pedestrians or bicyclists. To handle non-rigidly objects, Schuster et al. \cite{schuster2018sceneflowfields} builds the concept of sparse matching and uses an edge-preserving interpolation method \cite{revaud2015epicflow} to spread the sparse matches into the entire image. The reliability of generating these matches against light change, perspective deformations and occlusions is increased more by considering multiple frames of stereo \cite{schuster2019SceneFlowFieldsPlusPlus} and using a robust interpolation method \cite{hu2017robust}. The existing stereo-based methods achieve impressive results for seeking correspondences across the images. However, they are not reliable with matching textureless objects or challenging illumination. 
	
	Alternatively, some methods use active sensors for scene flow estimation. 
	All approaches relying on RGB-D cameras -- such as Kinect \cite{yoshida2017time} -- provide commendable results \cite{herbst2013rgb,quiroga2014dense,jaimez2015primal} but perform poorly in outdoor scenarios with large distances.
	Thus, LiDAR approaches were used mainly for our use-case.
	Dewan et al. \cite{dewan2016rigid} associate correspondences across LiDAR scans on point cloud space. They formulate the scene flow estimation as an energy minimization problem based on matching signature of histograms SHOT feature descriptors \cite{tombari2010unique}. However, such descriptors can be subjected to errors with sensor noise and they do not overcome the sparse nature of LiDAR data. Following the use of LiDAR, learning-based solutions exist to predict matches from point cloud. Some algorithms train neural network models from
	unstructured LiDAR point clouds \cite{liu2018learning, behl2018pointflownet}. However, training from that domain is a challenging task. Occupancy grids are used to structure LiDAR data 
	\cite{ushani2017learning, ushani2018feature} for building network models for feature learning on 3D space. 
	Vaquero et al. \cite{vaquero2018hallucinating} estimate dense optical flow for automotive applications from LiDAR scans only, but do not estimate scene flow.
	They build dense ground truth of the optical flow from the imagery part and they facilitate structuring LiDAR measurements by an alignment on the image domain. 
	Overall, LiDAR-based solutions result in competitive accuracy compared to stereo and RGB-D approaches. 
	However, they are not generating dense estimation of scene flow such as image-based solutions (e.g. stereo systems).
	
	The fusion of a sparse LiDAR and the imagery part has not yet been investigated in the context of scene flow.
	However, this fusion was applied to other related components.
	Ma et al. \cite{ma2018sparse} propose a deep regression network for dense depth estimation using one single camera. They use RGB-D raw data to train their model due to the lack of large-space data set with dense LiDAR. 
	The fusion of LiDAR and stereo is suggested to speed up the disparity map computation for outdoor scenes \cite{maddern2016real, huber2011integrating}. They interpolate LiDAR measurements on the image domain and utilize them as a prior information for computing the space of stereo disparity map. Park at el. \cite{park2018high} utilize deep convolutional neural network (CNN) architecture to fuse sparse LiDAR into dense stereo. Our work is the first among the aforementioned fusion methods to fuse a sparse LiDAR and stereo images for computing dense scene flow. 
	\newline

	\section{THE PROPOSED APPROACH} \label{Section2:LIDAR-FLOW APPROACH}
		For scene flow estimation, we require consecutive scans at times $t_0$ and $t_1$ from synchronized LiDAR and stereo frames $ \left(D_l^0, D_l^1, I_l^0, I_l^1, I_r^0, I_r^1 \right)$, where $D$ denotes sparse LiDAR measurements and $\left(I_l, I_r\right)$ are the left and right of the stereo images (see Fig. \ref{Fig2:Overview}).
		We assume the LiDAR measurements and left images are aligned (i.e. all sensors are calibrated). In other words, $D_l^0$ and $D_l^1$ are aligned to the images of $I_l^0$ and $I_l^1$ respectively. 	
		The stereo images are rectified and the baseline $B$ is known.
		
		Scene flow is represented by the optical flow components ${\left(u,v\right)}^T$ and the disparity values ${\left(d_0, d_1\right)}^T$ at a given pixel $p$ on image $I_l^0$ (i.e. the reference frame). 
		\begin{equation}
			SF(p) = {\left(u,v,d_0,d_1\right)}^T
			\label{SceneFlowVector}
		\end{equation}
		 We are denoting the image space information as well as LiDAR frames with capital letters and the pixel parameters with small letters.  
	
		\subsection{LiDAR-based Support Window} \label{SubSection1:LiDAR-basedSupportWindow}
			A prior smoothness assumption is made to improve the matching algorithm. A constant LiDAR measurement is used within a support window to facilitate the initialization and the propagation. 
			In detail, within one support window, $d_0$ is initialized on the coarsest level with the value of the closest LiDAR measurement of $D_l^0$ to support the propagation of robust LiDAR measurements into regions with less reliable image matching cost (i.e. minimum matching cost does not necessarily meet optimal scene flow in untextured regions). 
			Even with accurate optical flow components ${\left(u,v\right)}^T$, it is unlikely to match a measurement of $D_l^1$ because of the sparsity. This problem can be solved using the proposed support window also.  
			\begin{figure}[t]
				\centering
				\includegraphics[scale=1.0, width = 0.45\textwidth]{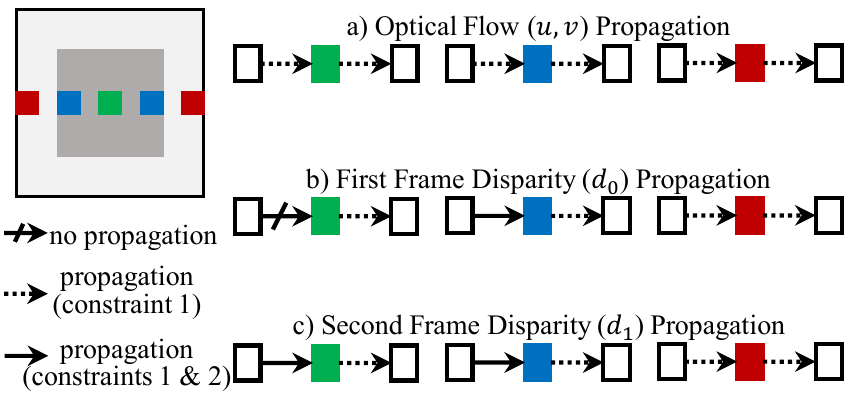}
				\caption{Our constrained propagation: Pixel with LiDAR measurement (green), pixels within LiDAR-based support window (blue) and image-based matches (red). Propagating into pixels should minimize matching cost (constraint 1) and yield geometry matches within the disparity space (constraint 2) defined by LiDAR measurement. Propagation of ${\left(u,v\right)}$ into all pixels as well as $d_0$ and $d_1$ into red pixels is applied only by constraint 1. Propagating $d_1$ into green and blue pixels as well as $d_0$ into blue pixels follows two constraints 1 \& 2. Propagation of $d_0$ into green pixel is not admitted. }
				\label{Fig3:InitializationAndPropagation}
			\end{figure}			
			Increasing the size of the support window has a big advantages in occluded areas (i.e. not visible in the right view) as long as the windows do not overlap.
			However, the support window can loose its advantages if a strong depth discontinuity occurs within the support window.

 			To compensate this drawback, a consistency check for the geometry within the support window is applied. The reliable LiDAR measurements of $D_l^0$ are not subjected to the consistency check.
		
		\subsection{LiDAR-Supported Matching} \label{SubSection2:LiDAR-SupportedMatching}
			We follow the coarse-to-fine approach of \cite{hu2016efficient} for seeking the matches across the images. Starting with the coarsest level of the reference view, $d_0$ for each pixel will be initialized with the nearest LiDAR measurement. The proposed support window limits the search area. The other scene flow components are initialized with an image-based method. This is also used for $d_0$ if no LiDAR measurement is met. On each pyramid level, the initial matches will be propagated into neighboring pixels with random search for a fixed number of iterations. The results of the finest level yields our final matches.

			The initialization and the propagation algorithm follow \cite{schuster2018sceneflowfields} but with further constraints to account for the additional sparse LiDAR measurements.
			
			Since we are considering LiDAR of $D_l^0$ as exact, propagation of $d_0$ into LiDAR seeds will not be allowed. Propagation from LiDAR seeds into the neighbor pixels within the LiDAR support window will be possible under a constraint (called constraint 2). This constraint reduces the possibilities of the disparity values coming from propagation or random search. Fig. \ref{Fig3:InitializationAndPropagation} illustrates the initialization and the propagation constraints in our algorithm. 
			Optimal scene flow matching is obtained by minimizing the matching cost (called constraint 1) for each pixel. The cost function is the sum of Euclidean norms between SIFT Flow features \cite{liu2011sift} over a patch window. If pixel $p_A$ in image $I_A$ matches $p_B$ in image $I_B$, then matching cost yields:
			\begin{equation}
				\begin{aligned}
				E(I_A, I_B, p_A, p_B) = \sum_{\Omega} 	
				\Vert 
				\phi \left(I_A \left(p_A \right)\right) - 
				\phi \left(I_B \left(p_B \right)\right)  
				\Vert
				\label{Eq: MatchingCost}
				\end{aligned}	
			\end{equation}
			Where $\Omega$ is 7x7 patch window and $\phi \left(I(p)\right)$ is the SIFT Flow vector of pixel $p$ in image $I$.
			Depending on whether a LiDAR measurement in $D_l^0$ or $D_l^1$ is hit during matching, four different cases for the matching are considered.
		    \begin{figure}[t]
		    	\centering
		    	\includegraphics[scale=1.0, width = 0.49\textwidth]{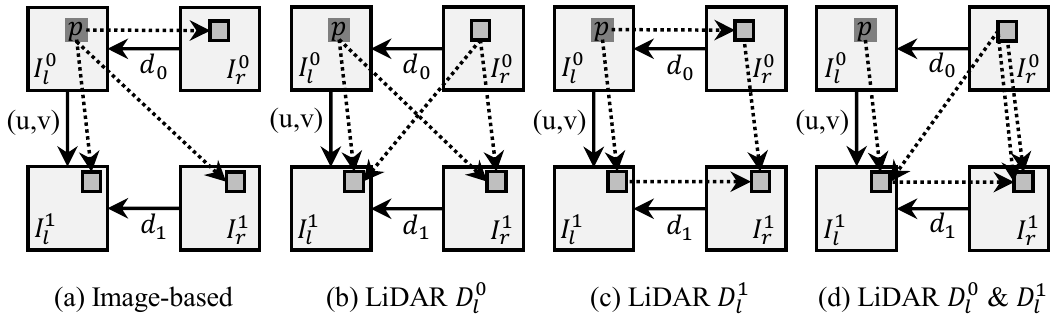}
		    	\caption{Matching cost computation: The scheme of pure image-based constancy (a), the constancy if a LiDAR measurement in $D_l^0$ is met (b), if a LiDAR measurement in $D_l^1$ is met (c) and if both LiDAR $D_l^0$ \& $D_l^1$ measurements are met (d).}
		    	\label{Fig4:MatchingCost}
		    \end{figure}
			The first case is purely image-based, i.e. no LiDAR measurement:
			\begin{equation}
			\begin{aligned}
			C_{I} = E_{I_l^0,I_l^1} + E_{I_l^0, I_r^0} + E_{I_l^0, I_r^1}
			\label{Eq: StereoMatching}
			\end{aligned}	
			\end{equation}
			where
			\begin{equation}
			\begin{aligned}
			E_{I_l^0,I_l^1} = E\left(I_l^0, I_l^1, p, p + (u, v) ^ T\right),
			\label{OpticalFlowConstancyOfLeft_Eq}
			\end{aligned}	
			\end{equation}
			
			\begin{equation}
			\begin{aligned}
			E_{I_l^0, I_r^0} = E\left(I_l^0, I_r^0, p, p - (d_0, 0) ^ T\right), 	
			\label{StereoConstancyOfT0_Eq}
			\end{aligned}	
			\end{equation}

			\begin{equation}
			\begin{aligned}			
			E_{I_l^0, I_r^1} = E\left(I_l^0, I_r^1, p, p + (u - d_1, v) ^ T\right).
			\label{CrossConstancyOfL0R1_Eq}
			\end{aligned}	
			\end{equation}	
			
			The second case is applied when a LiDAR measurement is available in $D_l^0$.
			In that case, the term $E_{I_l^0, I_r^0}$ in Eq. \ref{Eq: StereoMatching} will be replaced with two other constancy terms:
			
			\begin{equation}
			\begin{aligned}
			C_{D_l^0} = E_{I_l^0,I_l^1} + E_{I_l^0, I_r^1} + E_{I_r^0, I_r^1} + E_{I_r^0, I_l^1}
			\label{LiDAR_D0_Eq}
			\end{aligned}	
			\end{equation}
			
			where
			\begin{equation}
			\begin{aligned}
			E_{I_r^0, I_r^1} = E\left(I_r^0, I_r^1, p - (d_0, 0) ^ T, p + (u - d_1, v) ^ T\right),
			\label{OpticalFlowConstancyOfRight_Eq}
			\end{aligned}	
			\end{equation}
				
			\begin{equation}
			\begin{aligned}
			E_{I_r^0, I_l^1} = E\left(I_r^0, I_l^1, p - (d_0, 0) ^ T, p + (u, v) ^ T\right). 
			\label{CrossConstancyOfR0L1_Eq}
			\end{aligned}	
			\end{equation}
			
			The third case covers the availability of a LiDAR measurement in $D_l^1$. Two other constancy terms replace the term of $E_{I_l^0, I_r^1}$ in Eq. \ref{Eq: StereoMatching}: 
			\begin{equation}
			\begin{aligned}
			C_{D_l^1} = E_{I_l^0,I_l^1} + E_{I_l^0, I_r^0} + E_{I_r^0, I_r^1} + E_{I_l^1, I_r^1}
			\label{LiDAR_D1_Eq}
			\end{aligned}	
			\end{equation}
			\begin{equation}
				\begin{aligned}
				E_{I_l^1, I_r^1} = E\left(I_l^1, I_r^1, p + (u, v) ^ T, p + (u - d_1, v) ^ T\right) 
				\label{StereoConstancyOfT1_Eq}
				\end{aligned}	
			\end{equation}
			The last case involves LiDAR measurements in both frames $D_l^0$ and $D_l^1$:
			\begin{equation}
			\begin{aligned}
				C_{D_l^0D_l^1} = E_{I_l^0,I_l^1} + E_{I_r^0, I_l^1} + 2\cdot E_{I_r^0, I_r^1} + E_{I_l^1, I_r^1}
				\label{LiDAR_D0D1_Eq}
			\end{aligned}	
			\end{equation}
			The optical flow constancy of the right camera (Eq. \ref{OpticalFlowConstancyOfRight_Eq}) is used twice in this case, to avoid an improper deviation of the optical flow to match sparse LiDAR measurements in $D_l^1$.
			The proposed schemes of the matching cost are visualized in Fig. \ref{Fig4:MatchingCost} and the overall process of matching is shown in Fig. \ref{Fig2:Overview}.
					
			\begin{figure}[t]
				\centering
				\includegraphics[scale=1.0]{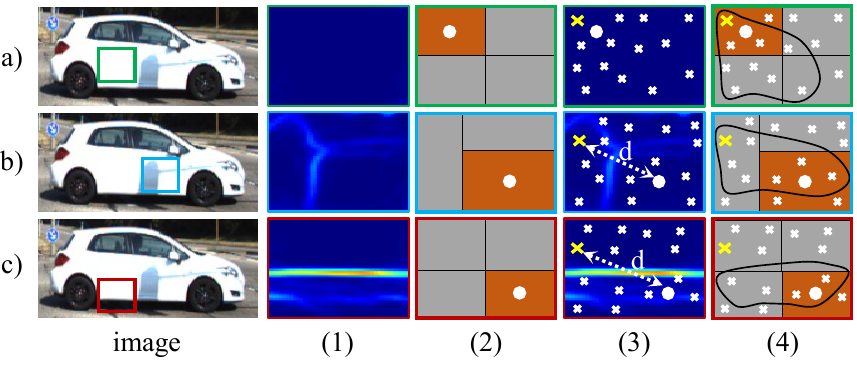}
				\caption{Constraints of the piecewise model computation for a superpixel (orange); whose center (solid white) in different regions (green, blue and red) in an image. Edge map (1), associated superpixels (e.g. 4 segments) to each region (2), associated matches to each region (crosses) (3) and consistent matches within the sketch (4). LiDAR match (yellow cross) initializes a piecewise model and assists in selecting more support matches for a final model (a), the same if the distance d from LiDAR match (yellow cross) to superpixel center is less than a defined threshold (b), otherwise a closest pure image-based match (closest white cross) is the best choice for initialization and estimating consistent matches (c). }
				\label{Fig5:Interpolation}
			\end{figure}	
			 
		\subsection{LiDAR-Supported Consistency Check} \label{SubSection3:LiDAR-Supported Consistency Check}
			Since we have sparse LiDAR measurements not covering the entire image, a consistency check is required to remove mismatches caused by image matching. Removing outliers is performed through two stages, each followed by a clustering algorithm. During all filtering stages sparse LiDAR measurements will never be filtered. Moreover, LiDAR measurements are employed as reference values during the clustering algorithm.
			
			The first stage involves checking the consistency of the disparity at $t_0$ (see Fig. \ref{Fig2:Overview}). To this end, the $d_0$ values of the scene flow matching are compared to a disparity map of $(I_l^0, I_r^0)$ computed with SGM \cite{hirschmuller2008stereo}. Values within the support windows are middle rank confident values and will not be subjected to filtering in this first stage.
			
			The second stage compares the scene flow matches using a forward-backward scene flow consistency check (see Fig. \ref{Fig2:Overview}). In the backward scene flow, the correspondences will be computed with respect to $I_l^1$ by the matching framework proposed in the previous section. Scene flow matches that yield differences bigger than a defined threshold will be removed. During the second stage, scene flow matches within the support window, as well as the components ($u,v,d_1$) of LiDAR pixels, are also subjected to filtering.
			
			The clustering algorithm segments the remaining scene flow matches into similar regions. The scene flow of LiDAR pixels will be used as reference values in the clustering algorithm. Small clusters which are having less pixels than a given threshold, will be removed. 
			
		\subsection{LiDAR-Supported Robust Interpolation}
		\label{Subsection4:LiDAR-SupportedRobustInterpolation}
		The proposed consistency check algorithm generates a sparse set of matches.
		To speed up the interpolation, the remaining scene flow pixels, except LiDAR measurements, will be subjected to a sparsification process as in \cite{schuster2018sceneflowfields} (see Fig. \ref{Fig2:Overview}). It selects within non-overlapping $3\times3$ blocks the scene flow match with the highest consistency.	
		After sparsification, we need to interpolate the remaining matches across the entire image into a dense scene flow (see Fig. \ref{Fig2:Overview}).
		The geometry and motion matches of LiDAR pixels will initialize piecewise models of the edge-preserving interpolation approach \cite{hu2017robust} during which the remaining outliers due to image matching are rejected. 	
		The reference view image $I_l^0$ is over-segmented into a set of superpixels and edges. The initial anchor point for each superpixel is its center of mass.
		Each superpixel is associated with the closest LiDAR measurement and LiDAR match for interpolation of geometry and motion respectively. If the closest LiDAR point lies within the superpixel or is in its proximity (defined by a certain threshold), it will be used as new anchor point for this superpixel. The edge-aware neighborhoods for each superpixel are computed with respect to its anchor point. A geodesic distance transformation, which is built with respect to an input edge map, assists in finding the edge-aware neighborhoods. We use the result of the Structured Edge Detector framework   \cite{dollar2013structured} to compute our edge map of the reference image. Fig. \ref{Fig5:Interpolation} visualizes the use of LiDAR matches for interpolation.		
			
		Based on the anchor points and their neighborhoods, each superpixel will be initialized with piecewise models for superpixel flow. These models are refined with random search and propagation between superpixels. During model refinement, hypotheses for geometry and motion are directly rejected if they are inconsistent with the associated LiDAR measurement or match.
				
	\section{EXPERIMENTS AND EVALUATION}
		We perform a series of experiments to verify the results of our new \name{} algorithm. First, we quantify the impact of fusing a sparse LiDAR into each step of the proposed pipeline (overview in Fig. \ref{Fig2:Overview}). Then, we follow with quantitative and qualitative comparison against a state-of-the-art image-only approach. Finally, the impact of using more samples of LiDAR measurements is evaluated. 

		\subsection{Evaluation Data}  
		\textbf{Evaluation Data}: 
		The first aim of our evaluation is emphasizing the capability of \name{} under challenging conditions. To the best of our knowledge, KITTI  \cite{menze2015object,geigerwe} is the only data set that provides stereo images in a combination with LiDAR measurements for real traffic scenarios. The train set of KITTI-2015 consists of 200 consecutive frames; each with ground truth of the scene flow components $(u,v,d_0, d_1)$. The ground truth has been generated by synchronizing the stereo with a high dense LiDAR and the augmentation with 3D CAD models for all vehicles in motions \cite{menze2015object}. 
		The evaluation on KITTI computes the average of percentage pixels whose the endpoint error of at least one scene flow component exceeds 3 pixels or 5\%. 
		The results of scene flow are quantified as an average over all frames in terms of endpoint error and outlier rate for the disparity maps $(D_0, D_1)$, the optical flow ($Fl$) and the scene flow ($SF$). 
		\begin{figure}[t]
			\centering
			\includegraphics[scale=1.0]{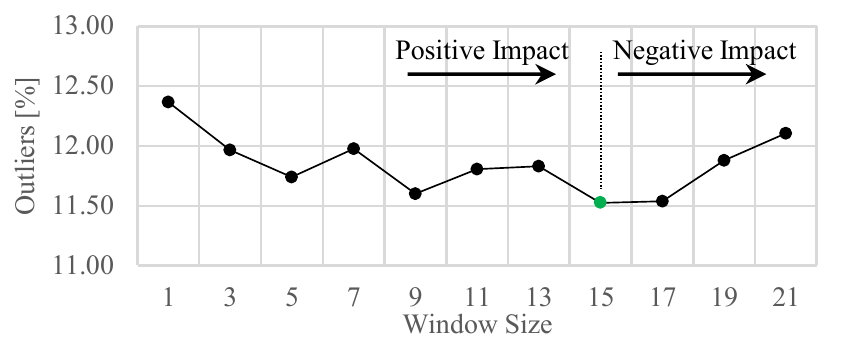}
			\caption{Impact of the window support size on the final scene flow outliers.}
			\label{Fig6:SupportWindowSize}
		\end{figure}
		
		\textbf{\name{} Input Data}: 
		To carry out our \name{} approach, we take the advantages of KITTI-2015 train set to build the input frames (mentioned in Section \ref{Section2:LIDAR-FLOW APPROACH}). Since the LiDAR measurements of KITTI are semi-dense and have been aligned into the reference view $I_l^0$, we have to preprocess the ground truth over all 200 frames. The preprocessing employs the optical flow ground truth to de-warp each geometry pixel of $D_l^1$ into $I_l^1$. That mimics the real capture of the second LiDAR frame. The de-warping process is supported with an occlusion-handling algorithm. Finally, we conduct a dense-to-sparse algorithm on each LiDAR frame $D_l^0$ and the de-warped $D_l^1$ to remain only 88 sparse LiDAR measurements (i.e. around 0.1\% of ground truth) as an average over all train set frames. Each sparse LiDAR map is generated by moving a $5\times5$ window with regular steps over the image and picking a LiDAR measurement within the defined window which is the nearest one to its center pixel.  
		After de-warping and sparsification, the correspondences of the two sparse measurements $(D_l^0,D_l^1)$ are completely dissolved in our input data. 
		\begin{figure*}[t]
			\centering
			\includegraphics[scale=1.0, width=\textwidth]{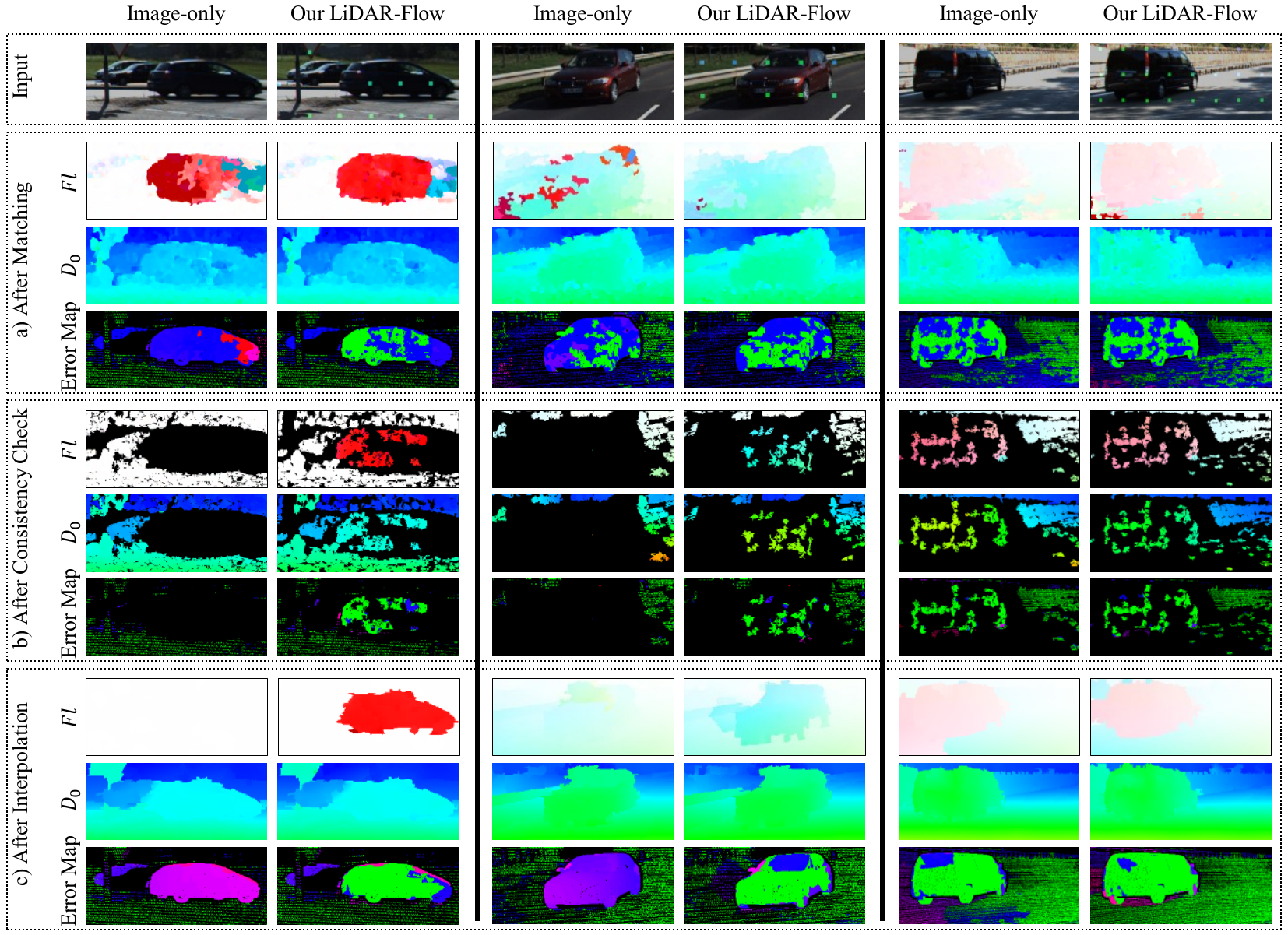}
			\caption{Three examples visualize the impact of our \name{} compared to an image-only SFF++ \cite{schuster2019SceneFlowFieldsPlusPlus}.
				Our \name{} performs superior in challenging image regions with shadows and low illumination. 
				Exactly these regions are of high importance for applications like autonomous driving or mobile robots.}
			\label{Fig8:ComparingToImag}
		\end{figure*}
	    \renewcommand{\arraystretch}{1.1}	
	    \begin{table*}[t]
	    	\caption{Quantitative evaluation for each single processing in Fig. \ref{Fig2:Overview}, and comparison of \name{} against image-only SFF++ \cite{schuster2019SceneFlowFieldsPlusPlus} across the main steps: matching, consistency check and interpolation. 
	    	}
	    	\label{Table1:PipelineEvaluation}
	    	\begin{center}
	    		\begin{tabular}{l l| |c|c|c| |c|c|c|c| |c}
	    			& & \multicolumn{3}{c||}{EPE [px]} & \multicolumn{4}{c||}{Outliers [\%]} & $Density$\\	
	    			& & $D_0$ & $D_1$ & $Fl$ & $D_0$ & $D_1$ & $Fl$ & $SF$ & [\%]\\
	    			\hline \hline	
	    			\multirow{5}{*}{Matching} & im-only             & 7.30 & 10.98 & 38.64 & 12.48 & 28.86 & 32.02 & 39.09 & 100 \\ 
	    			& im+LiDAR            & 6.90 & 14.19 & 36.91 & 12.48 & 28.70 & 32.67 & 39.55 & 100 \\ 
	    			& im+LiDAR+win        & 6.48 & 12.20 & 37.72 & 11.66 & 27.95 & 32.15 & 38.89 & 100 \\ 
	    			& im+LiDAR+win+const1 & \textbf{6.56} & 10.65 & 35.11 & 11.67 & 27.62 & 31.82 & 38.62 & 100 \\
	    			& \textbf{\name{}}: im+LiDAR+win+const1+const2 & 6.57 & \textbf{10.61} & \textbf{34.81} & \textbf{11.65} & \textbf{27.59} & \textbf{31.70} & \textbf{38.46} & 100 \\ 
	    			\cline{2-10}		 
	    			& SFF++ \cite{schuster2019SceneFlowFieldsPlusPlus} & 7.30 & 11.14 & 38.02 & 12.48 & 28.82 & 31.95 & 39.04 & 100 \\  
	    			
	    			\hline	
	    			\multirow{3}{*}{Consist. Check} & Disp+LiDAR & 0.85 & 6.50 & 23.49 & 2.01 & 16.00 & 20.15 & 22.71 & 63.75 \\
	    			& \textbf{\name{}}: disp+LiDAR \& backward+LiDAR & \textbf{0.71} & \textbf{0.93} & 1.46 & \textbf{1.21} & \textbf{2.46} & 2.97 & \textbf{4.21} & 37.30 \\	
	    			\cline{2-10}
	    			& SFF++ \cite{schuster2019SceneFlowFieldsPlusPlus} & 0.89 & 1.01 & \textbf{1.11} & 2.66 & 3.39 & \textbf{2.37} & 5.04 & 40.89 \\
	    			
	    			\hline	  		
	    				
	    			\multirow{4}{*}{Interpolation} 
	    			& Spars+LiDAR & 0.75 & 1.36 & 2.98 & 1.50 & 3.69 & 4.40 & 6.01 & 5.78 \\
	    			& Spars+LiDAR \& Interp-only  & 1.17 & 1.69 & 5.05 & 5.30 & 8.55 & 10.32 & 12.27 & 100 \\
	    			\showrowcolors
	    			& \textbf{\name{}}: Spars+LiDAR \& Interp+LiDAR &  \textbf{1.14} & \textbf{1.62} & \textbf{4.46} & \textbf{5.02} & \textbf{8.07} & \textbf{9.61} & \textbf{11.52} & 100 \\
	    			\hiderowcolors
	    			\cline{2-10}
	    			& SFF++ \cite{schuster2019SceneFlowFieldsPlusPlus} & 1.33 & 1.88 & 5.32 & 6.44 & 9.67 & 12.24 & 14.51 & 100\\
	    		\end{tabular}
	    	\end{center}
	    \end{table*}
	    \begin{table}[t]
	    	\caption{Quantitative evaluation  of the final scene flow results for the LiDAR seeds, the $15\times15$ support window and overall pixels compared to SFF++ \cite{schuster2019SceneFlowFieldsPlusPlus}.}
	    	\label{Table2:SeedsComparison}
	    	\begin{center}
	    		\begin{tabular}{c| |c|c|c}
	    			& \multicolumn{3}{c}{SF Outliers [\%]} \\
	    			& Seeds & Window & Dense \\
	    			\hline \hline	
	    			
	    			SFF++ \cite{schuster2019SceneFlowFieldsPlusPlus} & 16.05 & 13.89 & 14.51 \\
	    			
	    			\textbf{\name{}} & \textbf{7.88} & \textbf{8.70} & \textbf{11.52} \\
	    			\hline
	    			Average no. of evaluated points & 88 & 6820 & 91875 \\			
	    		\end{tabular}
	    	\end{center}
	    \end{table}	
    
		\subsection{\name{} Pipeline Evaluation and Comparison to an Image-only Approach}	
			Table \ref{Table1:PipelineEvaluation} quantifies the results after each fusion step in our \name{} pipeline. Since \name{} incorporates LiDAR measurements and stereo images into dense scene flow estimation on image domain; \cite{dewan2016rigid, liu2018learning, behl2018pointflownet} can not be conducted to our evaluation. They exploit the full resolution of LiDAR on point cloud domain and they remove LiDAR points on grounds which is inconsistent to our algorithm. As mentioned in related work, there is no directly related algorithm utilizing a sparse LiDAR and stereo images for scene flow computation. 
			Thus, we compare to the image-only scene flow algorithm SFF++ \cite{schuster2019SceneFlowFieldsPlusPlus} in its dual-frame version. We compare against SFF++ \cite{schuster2019SceneFlowFieldsPlusPlus}, since it follows a similar concept of sparse-to-dense interpolation; although it does not utilize any LiDAR data.
			The comparison to SFF++ \cite{schuster2019SceneFlowFieldsPlusPlus} involves the main steps of matching, consistency check and interpolation in Sections \ref{SubSection2:LiDAR-SupportedMatching}, \ref{SubSection3:LiDAR-Supported Consistency Check} and \ref{Subsection4:LiDAR-SupportedRobustInterpolation} respectively. 
			Fig. \ref{Fig8:ComparingToImag} visualizes also the  superior performance of our \name{} against SFF++ \cite{schuster2019SceneFlowFieldsPlusPlus} under three challenging scenarios in terms of ($FL$), ($D_0$) and error map which maps the inliers of scene flow with green color. 
			
			\textbf{Support Window}: Various sizes of the proposed support window are tested to check the impact into the final accuracy. We assume a symmetrical support window in our investigation.
			Fig. \ref {Fig6:SupportWindowSize} illustrates that increasing the size performs better accuracy until $15\times15$. However, the use of bigger sizes can drop the final accuracy down for the reasons mentioned in Section \ref{SubSection1:LiDAR-basedSupportWindow}. The following results are performed using the $15\times15$ support window.
								 			
		 	\textbf{Matching}: 
	 		Matching between stereo images only results in outlier rate $39.09\%$. With pushing the LiDAR measurements directly into matching algorithm, we constraint the space of finding other matches of $(u,v,d1)$ on the untextured regions. However, we affect badly reaching the optimal flow components in neighbor pixels during the propagation. One of the neighbor pixels within the proposed support window can reach the optimal matches with minimum matching cost. In addition, the use of matching cost constraint (const1) in Fig. \ref{Fig4:MatchingCost} and the disparity space constraint (const2) optimize the outlier rate to 38.46\%. The improvement is performed not only for the disparity map of $D_0$ but also for the matches of optical flow components $(Fl)$ and the disparity map of $(D_1)$.
	 		The quantitative values of our \name{} approach show improvement overall scene flow components compared to SFF++ \cite{schuster2019SceneFlowFieldsPlusPlus}. The error map shows in Fig. \ref{Fig8:ComparingToImag} more inliers with low illuminated regions of the cars, in which the matching algorithm of SFF++ \cite{schuster2019SceneFlowFieldsPlusPlus} fails. 
	
	 		\textbf{Consistency Check}: 
	 		The consistency check of the disparity map at $t_0$ using sparse LiDAR measurements (disp+LiDAR) removes the inaccurate geometry matches in which no contribution of LiDAR measurements is. That decreases the density of scene flow to 63.75\%, but remains the reliable geometry matches. The forward-backward consistency (backward+LiDAR) removes unreliable matches of optical flow $(u,v)$ and disparity $d_1$. That decreases the density of matches to 37.30\%. Together, consistency check of LiDAR results in more inliers to the critical regions (e.g. untextured)  where consistency check of SFF++ \cite{schuster2019SceneFlowFieldsPlusPlus} fails (see Fig. \ref{Fig8:ComparingToImag}). The quantitative comparison compared to SFF++ \cite{schuster2019SceneFlowFieldsPlusPlus} shows that less outlier rate of scene flow $(SF)$ and disparity maps $(D_0,D_1)$ but not with $(Fl)$. The reason of that, the contribution of LiDAR measurements in estimating the optical flow $(Fl)$ components is less than that with disparity maps $(D_0,D_1)$. 
	 				
	 		\textbf{Interpolation}: 
	 		The sparsification (spars+LiDAR) continues removing less confident scene flow components and decreases the density of scene flow results to 5.7\%. The interpolation before fusing LiDAR measurements (interp-only) is compared to the interpolation algorithm constrained by LiDAR measurements (interp+LiDAR). Overall components, \name{} interpolation algorithm achieves a better outlier rate (11.52 \%) than the rate of SFF++ \cite{schuster2019SceneFlowFieldsPlusPlus} of 14.51\%. Not only with the disparity ($D_0$), but even with the optical components $(Fl)$ and disparity ($D_1$), the use of 88 LiDAR measurements only, can resolve the challenging regions. Beside the quantitative comparison, Fig. \ref{Fig8:ComparingToImag} shows the substantial improvement under the challenging regions of shadowed areas and low illuminated areas compared to an image-only approach. 
			\begin{figure}[tpb]
				\centering
				\includegraphics[scale=1.0]{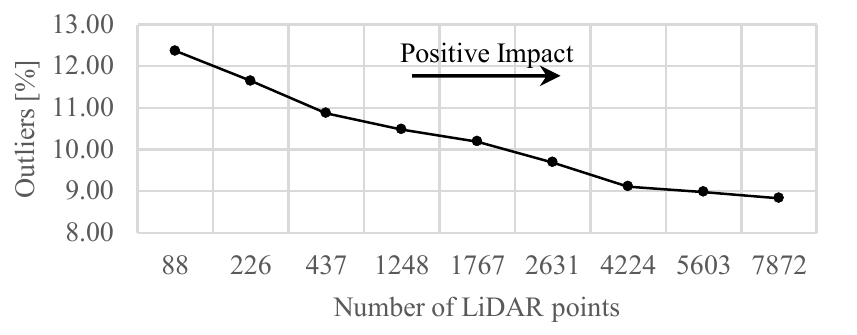}
				\caption{Impact of number of LiDAR measurements on the final scene flow outliers.}
				\label{Fig7:ResolutionTest}
			\end{figure}
  			In the aforementioned evaluation, error metrics are computed on average over the whole image.
  			Table \ref{Table2:SeedsComparison} indicates the final scene flow results in different image regions.
  			The results are given for the position of the 88 input LiDAR measurements only, indicating a halved outlier rate compared to the image-only approach.
  			In addition, the outliers in the $15\times15$ support window are substantially reduced.
  			This indicates that the utilized sparse LiDAR measurements serve as anchor points for these regions.
  			
  			\textbf{Run Time}: Estimating scene flow for a consecutive frame requires $78$ seconds on a single $4$ cores CPU $\MVAt$ $3.5$ GHz base frequency with \CC. Parallelism is utilized for a small part of the algorithm.
		\subsection{Impact of Number of LiDAR Measurements}
		We test the utilization of more LiDAR measurements and their impact on the final scene flow outlier rate. The proposed support window in Section \ref{SubSection1:LiDAR-basedSupportWindow} is neglected in order to avoid overlapping between LiDAR support windows. 
		Fig. \ref{Fig7:ResolutionTest} illustrates that the accuracy tends to be better with more LiDAR measurements. 
		However, even with very sparse LiDAR measurements (88 points only), we achieve already significant improvement.
		Such sparse LiDARs are comparably cheap and lead -- in combination with our \name{} -- to the desired increased robustness in challenging image regions.
		 
	 	\section{CONCLUSIONS}
	 	In this paper, we proposed \name{} to estimate dense scene flow from a fusion of sparse LiDAR and stereo images for resolving the challenges of textureless objects, shadows and poor illumination. 
	 	We tightly integrated highly accurate LiDAR measurements into each processing step of our pipeline. 
	 	They initialized image-based matching as well as served as anchor points for the consistency check in removing mismatches. 
	 	In addition, we incorporated the sparse LiDAR measurements as constraints to increase the robustness of the sparse-to-dense interpolation.
	 	In our comprehensive evaluation on KITTI, we demonstrated the impact of each design decision of our \name{}.
	 	Compared to image-only approaches, \name{} resolved robustly the lack of information in challenging image regions and improves the overall accuracy of scene flow. 
	 	Especially in critical scenarios -- see Fig. \ref{Fig8:ComparingToImag} with oncoming cars, low illumination, shadows -- our \name{} performed superior.


\bibliographystyle{IEEEtran}
\bibliography{IEEEabrv,LiDARFlowIROS}

\begin{thebibliography}{10}
\providecommand{\url}[1]{#1}
\csname url@rmstyle\endcsname
\providecommand{\newblock}{\relax}
\providecommand{\bibinfo}[2]{#2}
\providecommand\BIBentrySTDinterwordspacing{\spaceskip=0pt\relax}
\providecommand\BIBentryALTinterwordstretchfactor{4}
\providecommand\BIBentryALTinterwordspacing{\spaceskip=\fontdimen2\font plus
\BIBentryALTinterwordstretchfactor\fontdimen3\font minus
  \fontdimen4\font\relax}
\providecommand\BIBforeignlanguage[2]{{%
\expandafter\ifx\csname l@#1\endcsname\relax
\typeout{** WARNING: IEEEtran.bst: No hyphenation pattern has been}%
\typeout{** loaded for the language `#1'. Using the pattern for}%
\typeout{** the default language instead.}%
\else
\language=\csname l@#1\endcsname
\fi
#2}}

\bibitem{menze2015object}
M.~Menze and A.~Geiger, ``Object scene flow for autonomous vehicles,'' in
  \emph{IEEE International Conference on Computer Vision and Pattern
  Recognition (CVPR)}, 2015.

\bibitem{vcech2011scene}
J.~{\v{C}}ech, J.~Sanchez-Riera, and R.~Horaud, ``Scene flow estimation by
  growing correspondence seeds,'' in \emph{IEEE International Conference on
  Computer Vision and Pattern Recognition (CVPR)}, 2011.

\bibitem{wedel2011stereoscopic}
A.~Wedel, T.~Brox, T.~Vaudrey, C.~Rabe, U.~Franke, and D.~Cremers,
  ``Stereoscopic scene flow computation for 3d motion understanding,''
  \emph{International Journal of Computer Vision (IJCV)}, 2011.

\bibitem{thomas2019delio}
Q.~M. Thomas, O.~Wasenm{\"u}ller, and D.~Stricker, ``Delio: Decoupled lidar
  odometry,'' in \emph{IEEE International Conference on Intelligent Vehicles
  Symposium (IV)}, 2019.

\bibitem{schuster2019SceneFlowFieldsPlusPlus}
R.~Schuster, O.~Wasenm{\"u}ller, C.~Unger, G.~Kuschk, and D.~Stricker,
  ``{SceneFlowFields++}: Multi-frame matching, visibility prediction, and
  robust interpolation for scene flow estimation,'' \emph{arXiv preprint
  arXiv:1902.10099}, 2019.

\bibitem{hu2017robust}
Y.~Hu, Y.~Li, and R.~Song, ``Robust interpolation of correspondences for large
  displacement optical flow,'' in \emph{IEEE International Conference on
  Computer Vision and Pattern Recognition (CVPR)}, 2017.

\bibitem{hu2016efficient}
Y.~Hu, R.~Song, and Y.~Li, ``Efficient coarse-to-fine patchmatch for large
  displacement optical flow,'' in \emph{IEEE International Conference on
  Computer Vision and Pattern Recognition (CVPR)}, 2016.

\bibitem{schuster2018sceneflowfields}
R.~Schuster, O.~Wasenm{\"u}ller, G.~Kuschk, C.~Bailer, and D.~Stricker,
  ``Sceneflowfields: Dense interpolation of sparse scene flow
  correspondences,'' in \emph{IEEE Winter Conference on Applications of
  Computer Vision (WACV)}, 2018.

\bibitem{vedula1999three}
S.~Vedula, S.~Baker, P.~Rander, R.~Collins, and T.~Kanade, ``Three-dimensional
  scene flow,'' in \emph{IEEE International Conference on Computer Vision
  (ICCV)}, 1999.

\bibitem{wedel2008efficient}
A.~Wedel, C.~Rabe, T.~Vaudrey, T.~Brox, U.~Franke, and D.~Cremers, ``Efficient
  dense scene flow from sparse or dense stereo data,'' in \emph{European
  Conference on Computer Vision (ECCV)}, 2008.

\bibitem{li2008multi}
R.~Li and S.~Sclaroff, ``Multi-scale 3d scene flow from binocular stereo
  sequences,'' \emph{Computer Vision and Image Understanding (CVIU)}, 2008.

\bibitem{basha2013multi}
T.~Basha, Y.~Moses, and N.~Kiryati, ``Multi-view scene flow estimation: A view
  centered variational approach,'' \emph{International Journal of Computer
  Vision (IJCV)}, 2013.

\bibitem{hung2013consistent}
C.~H. Hung, L.~Xu, and J.~Jia, ``Consistent binocular depth and scene flow with
  chained temporal profiles,'' \emph{International Journal of Computer Vision
  (IJCV)}, 2013.

\bibitem{huguet2007variational}
F.~Huguet and F.~Devernay, ``A variational method for scene flow estimation
  from stereo sequences,'' in \emph{IEEE International Conference on Computer
  Vision (ICCV)}, 2007.

\bibitem{vogel20113d}
C.~Vogel, K.~Schindler, and S.~Roth, ``3d scene flow estimation with a rigid
  motion prior,'' in \emph{IEEE International Conference on Computer Vision
  (ICCV)}, 2011.

\bibitem{revaud2015epicflow}
J.~Revaud, P.~Weinzaepfel, Z.~Harchaoui, and C.~Schmid, ``Epicflow:
  Edge-preserving interpolation of correspondences for optical flow,'' in
  \emph{IEEE International conference on computer vision and pattern
  recognition (CVPR)}, 2015.

\bibitem{yoshida2017time}
T.~Yoshida, O.~Wasenm{\"u}ller, and D.~Stricker, ``Time-of-flight sensor depth
  enhancement for automotive exhaust gas,'' in \emph{IEEE International
  Conference on Image Processing (ICIP)}, 2017.

\bibitem{herbst2013rgb}
E.~Herbst, X.~Ren, and D.~Fox, ``Rgb-d flow: Dense 3-d motion estimation using
  color and depth,'' in \emph{IEEE International Conference on Robotics and
  Automation (ICRA)}, 2013.

\bibitem{quiroga2014dense}
J.~Quiroga, T.~Brox, F.~Devernay, and J.~Crowley, ``Dense semi-rigid scene flow
  estimation from rgbd images,'' in \emph{European Conference on Computer
  Vision (ECCV)}, 2014.

\bibitem{jaimez2015primal}
M.~Jaimez, M.~Souiai, J.~Gonzalez-Jimenez, and D.~Cremers, ``A primal-dual
  framework for real-time dense rgb-d scene flow,'' in \emph{IEEE International
  Conference on Robotics and Automation (ICRA)}, 2015.

\bibitem{dewan2016rigid}
A.~Dewan, T.~Caselitz, G.~D. Tipaldi, and W.~Burgard, ``Rigid scene flow for 3d
  lidar scans,'' in \emph{IEEE International Conference on Intelligent Robots
  and Systems (IROS)}, 2016.

\bibitem{tombari2010unique}
F.~Tombari, S.~Salti, and L.~Di~Stefano, ``Unique signatures of histograms for
  local surface description,'' in \emph{European Conference on Computer Vision
  (ECCV)}, 2010.

\bibitem{liu2018learning}
X.~Liu, C.~R. Qi, and L.~J. Guibas, ``Learning scene flow in 3d point clouds,''
  \emph{arXiv preprint arXiv:1806.01411}, 2018.

\bibitem{behl2018pointflownet}
A.~Behl, D.~Paschalidou, S.~Donn{\'e}, and A.~Geiger, ``Pointflownet: Learning
  representations for 3d scene flow estimation from point clouds,'' \emph{arXiv
  preprint arXiv:1806.02170}, 2018.

\bibitem{ushani2017learning}
A.~K. Ushani, R.~W. Wolcott, J.~M. Walls, and R.~M. Eustice, ``A learning
  approach for real-time temporal scene flow estimation from lidar data,'' in
  \emph{IEEE International Robotics Conference on Automation (ICRA)}, 2017.

\bibitem{ushani2018feature}
A.~K. Ushani and R.~M. Eustice, ``Feature learning for scene flow estimation
  from lidar,'' in \emph{Conference on Robot Learning (CoRL)}, 2018.

\bibitem{vaquero2018hallucinating}
V.~Vaquero, A.~Sanfeliu, and F.~Moreno-Noguer, ``Hallucinating dense optical
  flow from sparse lidar for autonomous vehicles,'' in \emph{IEEE International
  Conference on Pattern Recognition (ICPR)}, 2018.

\bibitem{ma2018sparse}
F.~Ma and S.~Karaman, ``Sparse-to-dense: Depth prediction from sparse depth
  samples and a single image,'' in \emph{IEEE International Conference on
  Robotics and Automation (ICRA)}, 2018.

\bibitem{maddern2016real}
W.~Maddern and P.~Newman, ``Real-time probabilistic fusion of sparse 3d lidar
  and dense stereo,'' in \emph{IEEE International Conference on Intelligent
  Robots and Systems (IROS)}, 2016.

\bibitem{huber2011integrating}
D.~Huber, T.~Kanade, \emph{et~al.}, ``Integrating lidar into stereo for fast
  and improved disparity computation,'' in \emph{IEEE International Conference
  on 3D Imaging, Modeling, Processing, Visualization and Transmission
  (3DIMPVT)}, 2011.

\bibitem{park2018high}
K.~Park, S.~Kim, and K.~Sohn, ``High-precision depth estimation with the 3d
  lidar and stereo fusion,'' in \emph{IEEE International Conference on Robotics
  and Automation (ICRA)}, 2018.

\bibitem{liu2011sift}
C.~Liu, J.~Yuen, and A.~Torralba, ``Sift flow: Dense correspondence across
  scenes and its applications,'' \emph{IEEE transactions on pattern analysis
  and machine intelligence (TPAMI)}, 2011.

\bibitem{hirschmuller2008stereo}
H.~Hirschmuller, ``Stereo processing by semiglobal matching and mutual
  information,'' \emph{IEEE Transactions on pattern analysis and machine
  intelligence (TPAMI)}, 2008.

\bibitem{dollar2013structured}
P.~Doll{\'a}r and C.~L. Zitnick, ``Structured forests for fast edge
  detection,'' in \emph{IEEE international conference on computer vision
  (ICCV)}, 2013.

\bibitem{geigerwe}
A.~Geiger, P.~Lenz, and R.~Urtasun, ``Are we ready for autonomous driving?'' in
  \emph{IEEE International Conference on Computer Vision and Pattern
  Recognition (CVPR)}, 2012.

\end{thebibliography}

\addtolength{\textheight}{-12cm}   

\end{document}